# Revisiting Aerial Scene Classification on the AID Benchmark

*Subhajeet Das[1], Susmita Ghosh[2], Abhiroop Chatterjee[2]*

[1] *Department of Computer Science and Engineering – AI, Brainware University, Kolkata, India*
[2] *Department of Computer Science and Engineering, Jadavpur University, Kolkata, India*

## ABSTRACT

**Aerial images play a vital role in urban planning and environmental preservation, as they consist of various structures, representing different types of buildings, forests, mountains, and unoccupied lands. Due to its heterogeneous nature, developing robust models for scene classification remains a challenge. In this study, we conduct a literature review of various machine learning methods for aerial image classification. Our survey covers a range of approaches from handcrafted features (e.g., SIFT, LBP) to traditional CNNs (e.g., VGG, GoogLeNet), and advanced deep hybrid networks. In this connection, we have also designed *Aerial-Y-Net*, a spatial attention-enhanced CNN with multi-scale feature fusion mechanism, which acts as an attention-based model and helps us to better understand the complexities of aerial images. Evaluated on the AID dataset, our model achieves 91.72% accuracy, outperforming several baseline architectures.**

## 1. INTRODUCTION

Aerial images serve a crucial role in revealing the Earth's surface and its structures, such as mountains, rivers, forests, residential areas, industrial regions, bridges, airports. Such images are captured from an elevated place by using Unmanned Aerial Vehicles (UAVs), satellites, aircraft, etc. Aerial images provide valuable insights into urban planning and development, disaster management, and environmental protection [1, 2]. However, dynamic imaging conditions pose a great challenge in classifying different objects on the Earth's surface due to image distortion, dissimilar illumination, and noise. Moreover, the intra-class and inter-class variability of the objects of the images makes the classification [1] process more challenging. Traditional aerial image classification methods heavily depend on manual interactions such as labeling, feature engineering, making the process prone to error, labor-intensive, and time-consuming. Automated technologies are required to speed up the classification process by minimizing human intervention. Several researchers have worked on classifying aerial images. This article provides an overview of various state-of-the-art approaches. In addition, a baseline feature extractor based on a convolutional neural network (CNN) has been designed for classification, employing a multi-branch feature fusion framework enhanced with an attention mechanism.

Rest of this article is organized as follows. Section 2 offers an analysis of prior works using several machine learning algorithms for classifying aerial images. Section 3 provides a framework of the proposed model, ***Aerial-Y-Net***. Section 4 outlines the dataset, hyper-parameters, evaluation metrics, and preprocessing strategies. Section 5 provides an analysis of results section, and Section 6 concludes the study with future directions.

## 2. LITERATURE REVIEW

Aerial image classification plays a pivotal role in determining land occupancy [3], urban planning [4], environmental preservation [5], etc. We list all the methods and their performance in Table **1**. **Xia et al.** introduced the *AID* [2] dataset and also evaluated the performance of various pre-trained networks on it. They divided the dataset into the non-overlapping train and test sets in a 5:5 ratio. They fine-tuned a pre-trained VGG-16 [6], GoogLeNet [7] and CaffeNet [8] that achieved an accuracy of 89.64%, 86.39%, and 89.53%, respectively, on the AID test set. In another work, **Bian et al.** [9] presented a saliency feature-based local binary pattern (salM$^3$LBP) and codebookless model (CLM) as the feature extractor for aerial image classification. They used the Kernel Extreme Learning Machine (KELM) [10] as the final classifier. With the combined salM$^3$LBP and CLM features, their approach obtained an overall accuracy of 89.76% on 50% test set of the AID data, and only for the salM$^3$LBP, the model achieved 87.59% accuracy. In [11], **Sun et al.** presented the Feature Pyramid Network (FPN) based method, where they used the VGG-19 model [6] as the backbone. They also considered Bag of Visual Words (BoVW) [12], Scale Invariant Feature Transform (SIFT), and Mean Squared Displacement (MSD) algorithm as the feature extractor. The combined BoVW and SIFT features, along with the pre-trained FPN, achieved 65.74% accuracy, BoVW and MSD features with FPN achieved 67.59% accuracy, whereas the standalone pre-trained VGG-19 model achieved 87.18% accuracy. The BoVW and VGG-19 FPN notably achieved 89.18% accuracy.

In [13], **Wei et al.** proposed a WGAN [14] based generative adversarial network, where the base architecture is modified in such a way that features of different levels are concatenated and fed into a 2-layered MLP for classification. Their proposed MF-WGAN model obtained 80.35% accuracy. **Wang et al.** [15] focused on extracting local semantic information and discriminative patterns using their MILRDA network that leverages CNN. The network uses Residual Dense Attention Block (RDAB) for extracting discriminative patterns while keeping the local patterns intact. The extracted patterns are broken down into a number of small patches, where the irrelevant patches are suppressed using channel attention-guided multi-instance adaptive subsampling layers. The MILRDA model has obtained 95.46% accuracy. **Wan et al.** [16] proposed a pre-trained ResNeXt [17] based lightweight channel attention enhanced feature fusion network named *LmNet*. Their channel attention block closely resembles the working principles of the Efficient Channel Attention (ECA) module [18]. However, instead of using an adaptive kernel size for the convolution operation in the attention block, they utilized fixed 1×1 kernels, and they used such a convolution layer twice in this block, one for contracting to fewer dimensions and another for expanding back to the original dimension. This gives an accuracy of 97.12%.

**Zhang et al.** [19] presented a transfer learning model that incorporates features extracted by pre-trained MobileNet V2 [20], which are further enhanced using dilated convolution for learning from the broader context and the Squeeze-and-Excitation (SE) [21] channel attention to guide the network to learn from relevant features. They have also introduced the multi-dilation-based subsampling layer that extracts and combines the features by first applying convolution kernels with different dilation rates and then performing the global average pooling operation to obtain the feature vector. Using this setup, their *SE-MDPMNet* model achieved an average accuracy of 97.14%. In another research, **Wang et al.** [22] proposed a method that automatically learns from the properties of the images. The high-resolution remote sensing images are first cropped into smaller regions, and then they are augmented to present different representations to the network. These images are fed to a Siamese network with varying levels of granularity that learns to differentiate between two distinct samples. The proposed canonical pooling layer selects the most informative representation by taking the element-wise maximum across intermediate outputs, effectively comparing and retaining the strongest signals from each feature. Finally, all these patterns are stacked and used for classification. The MG-CAP model achieves an accuracy of 96.12% on the AID dataset. In contrast, Cheng et al. [23] proposed the Discriminative CNN (D-CNN), which employs the pre-trained VGG-16 as its backbone and incorporates L2 regularization together with cross-entropy loss to reduce intra-class variance while enhancing inter-class separability. Using this strategy, the model reached an accuracy of 96.89%.

***Table 1:*** *Comprehensive Study of Different Methods. The highest result is marked in bold.*

| **Methods** | **Features** | **Accuracy (%)** |
|---|---|---|
| *VGG-16 [2, 6]* | *Pre-trained on ImageNet* | *89.64* |
| *GoogLeNet [2, 7]* | *Pre-trained on ImageNet* | *86.39* |
| *CaffeNet [2, 8]* | *Pre-trained on ImageNet* | *89.53* |
| *sal$M^3$LBP + CLM [9]* | *Local Binary Pattern, CLM* | *89.76* |
| *sal$M^3$LBP [9]* | *Local Binary Pattern* | *87.59* |
| *$M^3$LBP [9]* | *Local Binary Pattern* | *84.80* |
| *salCLM [9]* | *CNN Extracted* | *88.41* |
| *CLM [9]* | *CNN Extracted* | *87.33* |
| *BoVW + SIFT [11]* | *BoVW, SIFT* | *65.74* |
| *BoVW + MSD [11]* | *BoVW, MSD* | *67.59* |
| *VGG-19 (FPN) [11]* | *Pre-trained on ImageNet* | *83.46* |
| *VGG-19 (FC) [11]* | *Pre-trained on ImageNet* | *87.18* |
| *BoVW + FPN [11]* | *BoVW, CNN Extracted* | *89.18* |
| *SIFT + MSD + FPN [11]* | *SIFT, MSD, CNN Extracted* | *86.50* |
| *MF-WGAN [13]* | *CNN Extracted* | *80.35* |
| *MILRDA [14]* | *CNN Extracted* | *95.46* |
| *LmNet [16]* | *Pre-trained on ImageNet* | *97.12* |
| *SE-MDPMNet [19]* | *Pre-trained on ImageNet* | ***97.14*** |
| *MG-CAP [22]* | *CNN Extracted* | *96.12* |
| *D-CNN VGG-16 [23]* | *Pre-trained on ImageNet* | *96.89* |

## 3. METHODOLOGY

The article presents Aerial-Y-Net, a spatial attention-enhanced deep neural network for aerial image classification. It comprises two identical ARNets whose extracted features are concatenated and passed to a classifier. Additionally, a fusion spatial attention mechanism (FuSAM) is cascaded into two parallel branches of the main model. The complete pipeline is detailed below.

**Architecture of the ARNet Model.** The ARNet model contains four conv blocks with 64, 128, 256, and 512 kernels, where each of the blocks has a ReLU-activated convolution layer, one batch normalization layer, and one Max-pool layer with 2×2 kernel. The convolution layer extracts features, which are normalized using the batch norm layer, while the Max-pool operation downsamples the spatial dimension of the extracted feature maps. Each of these convolution blocks extracts features at different abstraction levels.

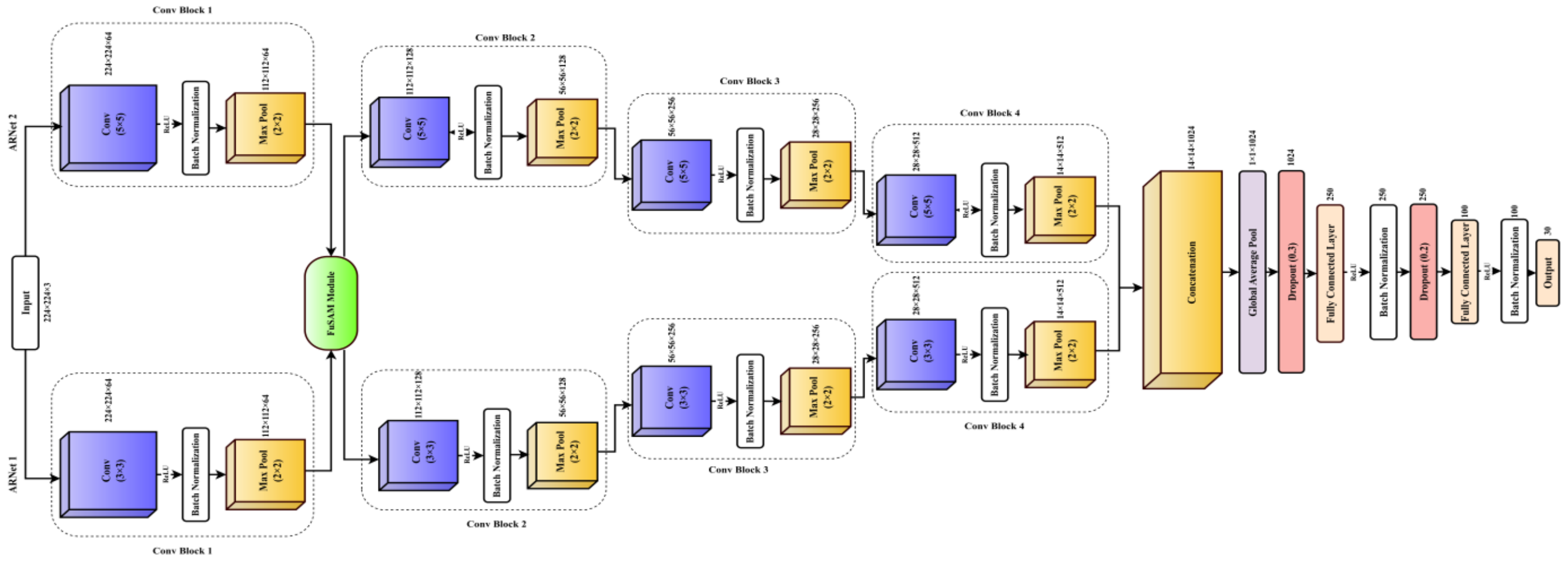

***Figure 1.*** *The Architecture of the Proposed Aerial-Y-Net Model.*

**The Architecture of the Aerial-Y-Net Model.** The Aerial-Y-Net architecture is built upon two parallel ARNet models, where the ARNet 1 model is initialized with a kernel size of 3×3 for all the convolution layers. Whereas, the ARNet 2 model is initialized with a convolution kernel size of 5×5. This setup helps the network to extract and learn from features at two distinct granularity levels.

ARNet 1 is designed to effectively extract local features, whereas ARNet 2, with its comparatively larger receptive field, focuses on capturing broader spatial information. The resulting feature maps from both networks are then concatenated along the channel dimension and passed through a Global Average Pooling (GAP) layer, producing a 1D feature vector. The vector is then fed to an MLP having 250 and 100 neurons with ReLU activation. Dropout [24] regularization with 30% and 20% dropout rates is applied before each of the FC layers, respectively, to reduce the risk of overfitting. The final layer comprises of 30 neurons with SoftMax activation for classification task.

**Fusion Spatial Attention Module.** The module, FuSAM, adaptively emphasizes the most prominent spatial regions of feature maps. This module in the Aerial-Y-Net bridges the two parallel networks and utilizes the feature maps derived from the initial convolution block of the two parallel ARNet models as its input and concatenates the feature vectors. Thereafter, this module performs a dilated convolution operation with a 3×3 kernel size, a dilation of 2, with sigmoid activation function, providing a spatial attention map, which is applied through the Hadamard product with the input feature maps for recalibration, as visualized in Figure **2**.

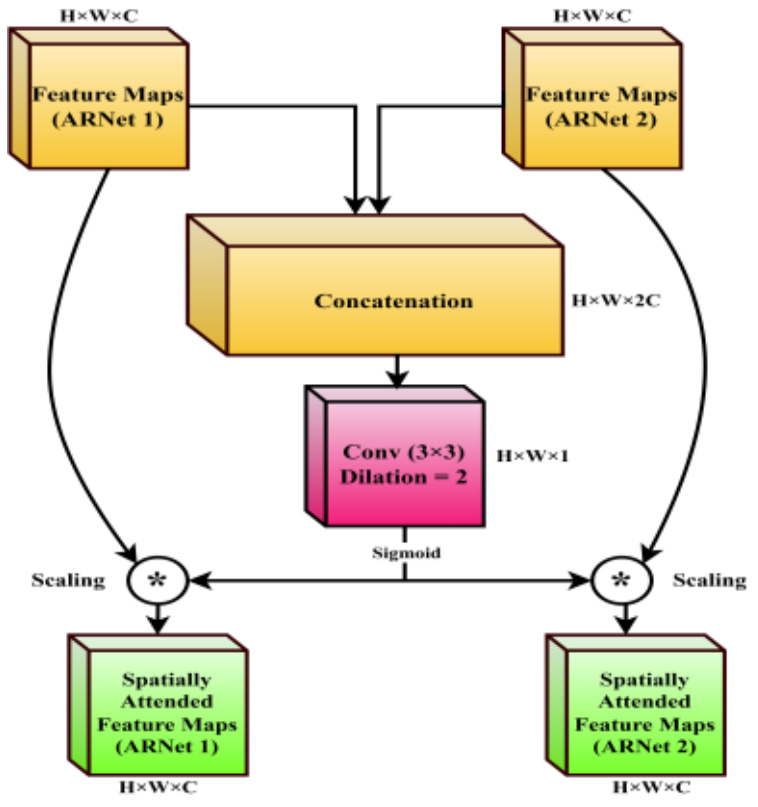

***Figure 2.*** *The Architecture of the FuSAM.*

## 4. EXPERIMENTAL SETUP

**Dataset.** As previously noted, the AID [2] dataset was employed, comprising a total of 10,000 aerial images obtained from Google Earth and categorized into 30 distinct classes. The dataset is divided into training and testing subsets using a 50:50 split. The class-wise distribution is illustrated in Figure 3.

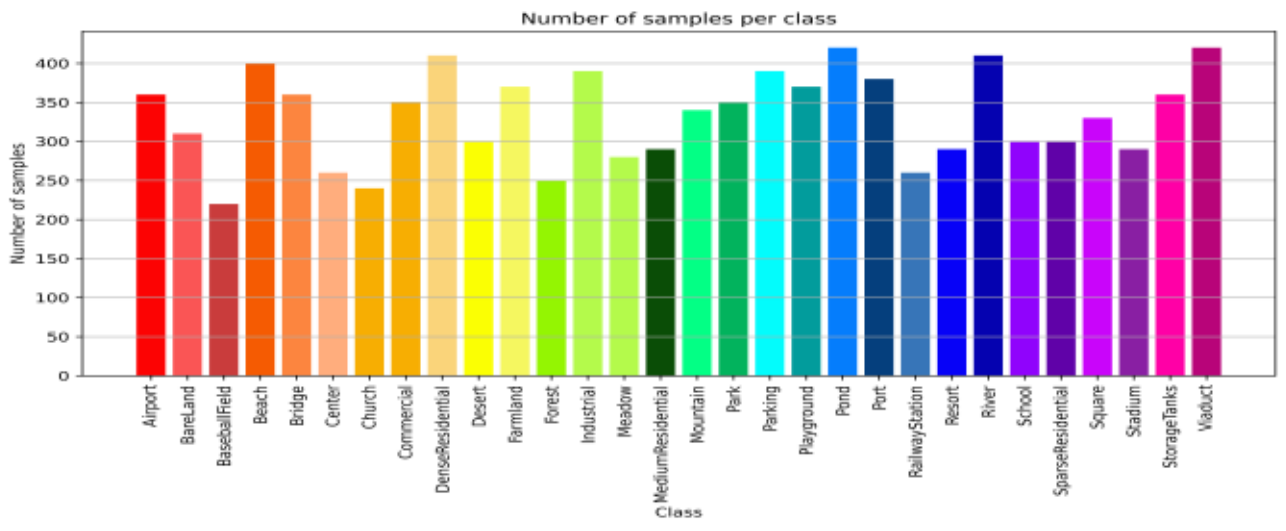

***Figure 3.*** *Class-wise Sample Distribution of the AID Benchmark Dataset.*

**Data Pre-processing.** The preprocessing includes resizing the images to dimensions 224×224×3. Then these are normalized by dividing each pixel intensity of every channel by 255 to scale them between the range [0, 1].

***Table 2.*** *Hyperparameters Used.*

| ***Parameters*** | ***Values*** |
|---|---|
| *Epochs* | *200* |
| *Optimizer* | *Adam* |
| *Learning Rate* | $1e^{-3}$ |
| *Learning Rate Scheduler* | *Cosine Annealing with Warm Restart* |
| *Scheduler Restart Frequency* | *50* |
| *Minimum LR for Scheduler* | $1e^{-6}$ |
| *Batch Size* | *32* |
| *Loss Function* | *Categorical Cross-Entropy* |

**Data Augmentation.** To address the data imbalance issue of the AID [2] dataset, we have used augmentation techniques like - horizontal & vertical flip with 50% probability, 90° rotation with 50% probability, brightness and contrast shift with 30% probability, RGB color shift with 50% probability, and median blurring with 40% probability on the training set. This introduces diversity into the samples of every batch and helps the model to generalize and learn better representations from the existing data.

**Hyperparameters Setup.** The initial learning rate of the proposed model is set to $1e^{-3}$, and gradually reduced to $1e^{-6}$ during the training using the Cosine Annealing with Warm Restart learning rate scheduling algorithm [24] with a restart frequency of 50 epochs. The training of the model is done for 200 epochs and a batch size of 32, as described in Table 2. Adam optimizer is used to optimize the model and scale the learning rate dynamically with respect to the model's parameters. Categorical cross-entropy loss is considered. The dropout regularization technique [25] is used to reduce the overfitting issue.

**Performance Metrics Used.** The model's classification performance is analyzed using metrics like accuracy, precision, recall, and the F1-score.

## 5. RESULTS ANALYSIS

The Aerial-Y-Net, while assessed on the test set of the AID dataset, achieved 91.72% accuracy, 91.80% precision, 91.72% recall, and 91.70% F1-Score. Table 3 presents the Class-wise performance. The accuracy for a few of the classes is exceptionally high, while for the other classes, it is comparatively low. The changes in the model's accuracy and loss on the test set over the span of the training epochs are visualized in Figure **4**. It is evident that the accuracy of the model gradually increases over epochs. Furthermore, both the t-SNE and the ROC curve have also been plotted in Figures **5** and **6,** respectively.

***Table 3:*** *Class-wise Classification Performance.*

| ***Class*** | ***Accuracy (%)*** | ***Precision (%)*** | ***Recall (%)*** | ***F1-score (%)*** |
|---|---|---|---|---|
| *Airport* | *91* | *90* | *91* | *90* |
| *Bare Land* | *95* | *91* | *95* | *93* |
| *Baseball Field* | *99* | *89* | *99* | *94* |
| *Beach* | *97* | *99* | *97* | *98* |
| *Bridge* | *92* | *95* | *92* | *94* |
| *Center* | *80* | *82* | *80* | *81* |
| *Church* | *88* | *76* | *88* | *82* |
| *Commercial* | *92* | *89* | *92* | *91* |
| *Dense Residential* | *92* | *96* | *92* | *94* |
| *Desert* | *91* | *97* | *91* | *94* |
| *Farmland* | *94* | *96* | *94* | *95* |
| *Forest* | *97* | *100* | *97* | *98* |
| *Industrial* | *83* | *88* | *83* | *86* |
| *Meadow* | *98* | *97* | *98* | *98* |
| *Medium Residential* | *91* | *97* | *91* | *94* |
| *Mountain* | *99* | *96* | *99* | *97* |
| *Park* | *88* | *85* | *88* | *86* |
| *Parking* | *99* | *96* | *99* | *97* |
| *Playground* | *95* | *93* | *95* | *94* |
| *Pond* | *96* | *94* | *96* | *95* |
| *Port* | *95* | *94* | *95* | *95* |
| *Railway Station* | *85* | *81* | *85* | *83* |
| *Resort* | *80* | *86* | *80* | *83* |
| *River* | *94* | *95* | *94* | *94* |
| *School* | *77* | *85* | *77* | *81* |
| *Sparse Residential* | *99* | *92* | *99* | *96* |
| *Square* | *81* | *83* | *81* | *82* |
| *Stadium* | *86* | *91* | *86* | *88* |
| *Storage Tanks* | *91* | *96* | *91* | *94* |
| *Viaduct* | *98* | *91* | *98* | *94* |

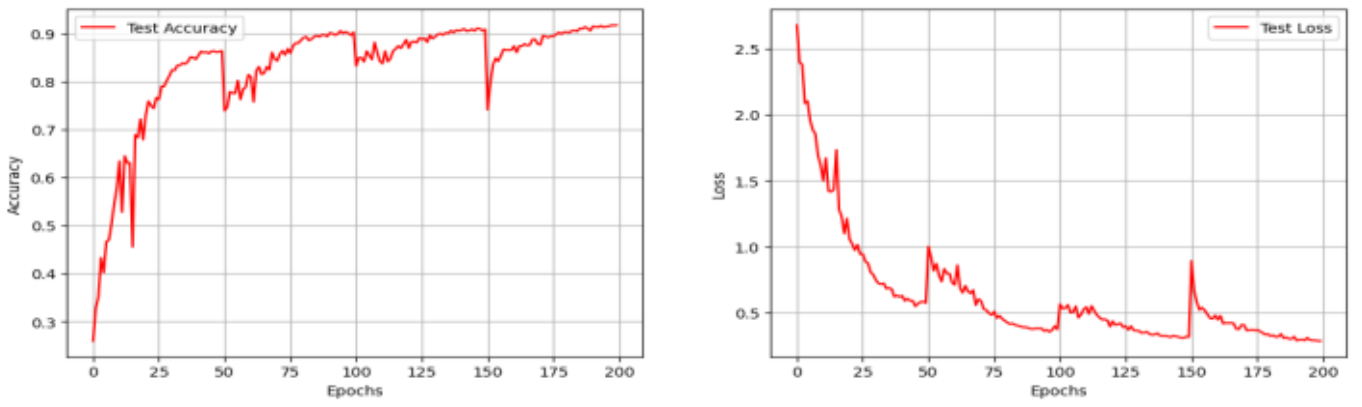

***Figure 4.*** *Accuracy and Loss over Epochs.*

The t-SNE (Figure **5**) shows overlapping clusters due to high inter-class similarity. This shows challenges in class-level separability despite strong overall accuracy.

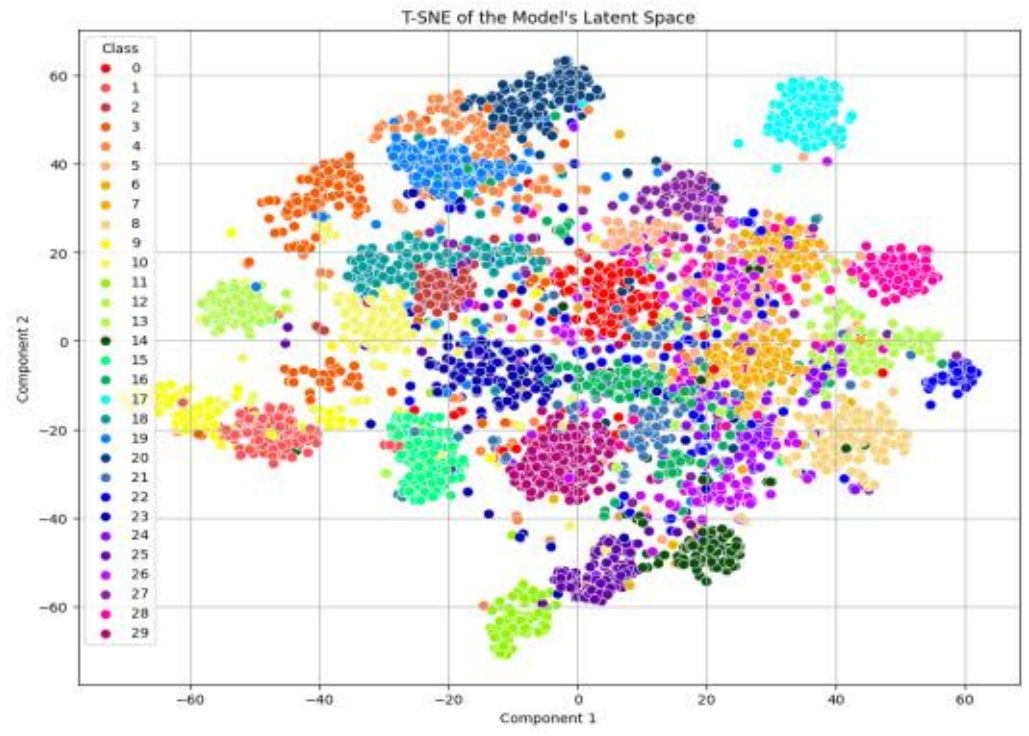

***Figure 5.*** *t-SNE of the Model on the Hold-Out Test Set.*

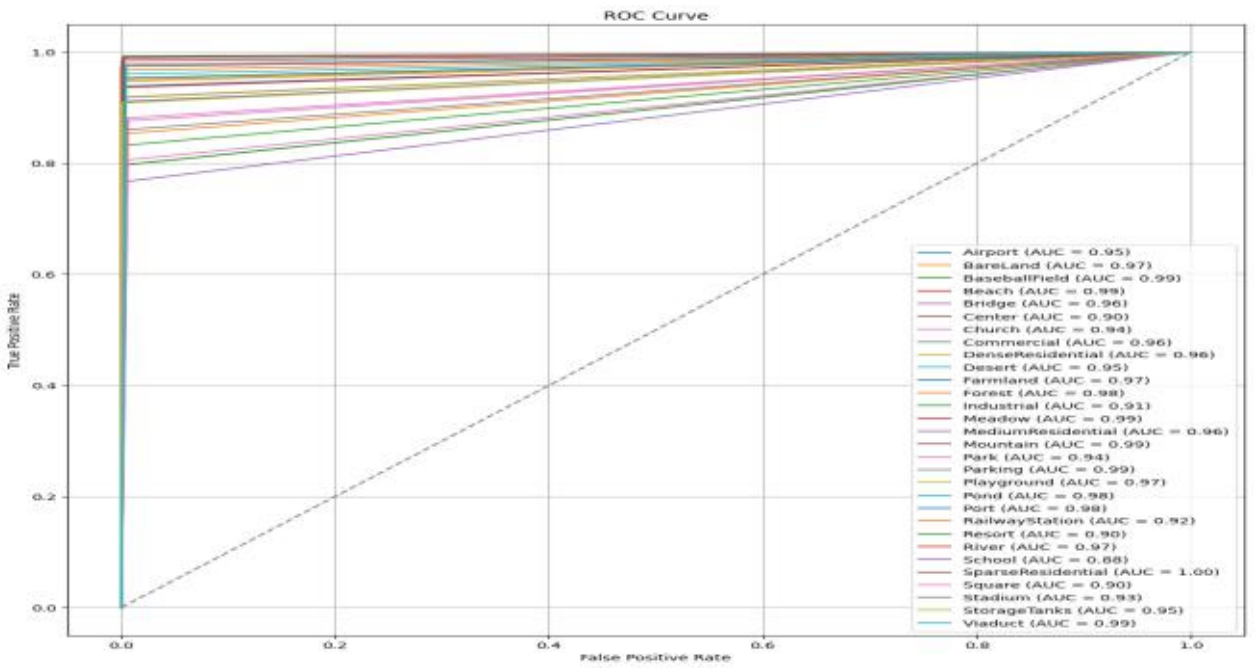

***Figure 6.*** *ROC Curve of the proposed model.*

## 6. CONCLUSION

The article presents a study of various methods to classify aerial images and also proposes the Aerial-Y-Net, a convolutional neural network-based spatial attention-enhanced feature fusion network for aerial image classification. To address the data imbalance issue, the method utilizes data augmentation techniques. To enhance the training accuracy, a dynamic learning rate scheduling algorithm is used, resulting in an accuracy of 91.72% on the test set of the AID dataset.

**Limitations and Future Works.** A key limitation is the model's reduced ability to distinguish visually similar classes with overlapping background features. The future work will include applying channel attention and self-attention mechanisms for efficient feature recalibration.

## 7. REFERENCES

[1] Y. Hua, L. Mou, and X. X. Zhu, "Relation Network for Multilabel Aerial Image Classification," *IEEE Transactions on Geoscience and Remote Sensing,* vol. 58, no. 7, pp. 4558 - 4572, 2020.

[2] G. S. Xia, J. Hu, F. Hu, B. Shi, X. Bai, and Y. Zhong, "AID: A Benchmark Data Set for Performance Evaluation of Aerial Scene Classification," *IEEE Transactions on Geoscience and Remote Sensing,* vol. 55, no. 7, pp. 3965 - 3981, 2017.

[3] Marcos, M. Volpi, B. Kellenberger, and D. Tuia, "Land Cover Mapping at Very High Resolution with Rotation Equivariant CNNs: Towards Small Yet Accurate Models," *ISPRS Journal of Photogrammetry and Remote Sensing,* vol. 145, pp. 96 - 107, 2018.

[4] N. Audebert, B. L. Saux, and S. Lefèvre, "Beyond RGB: Very High Resolution Urban Remote Sensing with Multimodal Deep Networks," *ISPRS Journal of Photogrammetry and Remote Sensing,* vol. 140, pp. 20 - 32, 2018.

[5] D. Wen, X. Huang, H. Liu, W. Liao, and L. Zhang, "Semantic Classification of Urban Trees Using Very High Resolution Satellite Imagery," *IEEE Journal of Selected Topics in Applied Earth Observations and Remote Sensing,* vol. 10, no. 4, pp. 1413 - 1424, 2017.

[6] K. Simonyan and A. Zisserman, "Very Deep Convolutional Networks for Large-Scale Image Recognition," *3rd International Conference on Learning Representations,* 2015.

[7] C. Szegedy, W. Liu, Y. Liu, P. Sermanet, S. Reed, D. Anguelov, D. Erhan, V. Vanhoucke, and A. Rabinovich, "Going Deeper with Convolutions," *Conference on Computer Vision and Pattern Recognition,* pp. 1 - 9, 2015.

[8] Y. Jia, E. Shelhamer, J. Donahue, S. Karayev, J. Long, R. Girshick, S. Guadarrama, and T. Darrell, "Caffe: Convolutional Architecture for Fast Feature Embedding," *22nd ACM International Conference on Multimedia,* pp. 675 - 678, 2014.

[9] X. Bian, C. Chen, L. Tian, and Q. Du, "Fusing Local and Global Features for High-Resolution Scene Classification," *IEEE Journal of Selected Topics in Applied Earth Observations and Remote Sensing,* vol. 10, no. 6, pp. 2889 - 2901, 2017.

[10] G. B. Huang, Q. Y. Zhu, and C. K. Siew, "Extreme Learning Machine: Theory and Applications," *Neurocomputing,* vol. 70, no. 1 - 3, pp. 489 - 501, 2006.

[11] X. Sun, Q. Zhu, and Q. Qin, "A Multi-Level Convolution Pyramid Semantic Fusion Framework for High-Resolution Remote Sensing Image Scene Classification and Annotation," *IEEE Access,* vol. 9, pp. 18195-18208, 2021.

[12] S. Gidaris, A. Bursuc, N. Komodakis, P. Perez, and M. Cord, "Learning Representations by Predicting Bags of Visual Words," *IEEE Conference on Computer Vision and Pattern Recognition,* pp. 6928 - 6938, 2020.

[13] Y. Wei, X. Luo, L. Hu, Y. Peng, and J. Feng, "An Improved Unsupervised Representation Learning Generative Adversarial Network for Remote Sensing Image Scene Classification," *Remote Sensing Letters,* vol. 11, no. 6, pp. 598 - 607, 2020.

[14] M. Arjovsky, S. Chintala, and L. Bottou, "Wasserstein Generative Adversarial Networks," *International Conference on Machine Learning, PMLR,* vol. 70, pp. 214 - 223, 2017.

[15] X. Wang, H. Xu, L. Yuan, W. Dai and X. Wen, "A Remote-Sensing Scene-Image Classification Method Based on Deep Multiple-Instance Learning with a Residual Dense Attention ConvNet," *Remote Sensing, MDPI,* vol. 14, no. 20, p. 5095, 2022.

[16] H. Y. Wan, J. Chen, Z. X. Huang, Y. Feng, Z. Zhou, X. P. Liu, B. D. Yao, and T. Xu, "Lightweight Channel Attention and Multiscale Feature Fusion Discrimination for Remote Sensing Scene Classification," *IEEE Access,* vol. 9, pp. 94586 - 94600, 2021.

[17] S. Xie, R. Girshick, P. Dollár, Z. Tu, and K. He, "Aggregated Residual Transformations for Deep Neural Networks," *IEEE Conference on Computer Vision and Pattern Recognition,* pp. 1492 - 1500, 2017.

[18] Q. Wang, B. Wu, P. Zhu, P. Li, W. Zuo et al, "ECA-Net: Efficient Channel Attention for Deep Convolutional Neural Networks," *Conference on Computer Vision and Pattern Recognition,* pp. 11534 - 11542, 2020.

[19] B. Zhang, Y. Zhang, and S. Wang, "A Lightweight and Discriminative Model for Remote Sensing Scene Classification with Multidilation Pooling Module," *IEEE Journal of Selected Topics in Applied Earth Observations and Remote Sensing,* vol. 12, no. 8, pp. 2636 - 2653, 2019.

[20] M. Sandler, A. Howard, M. Zhu, A. Zhmoginov, and L. C. Chen, "MobileNetV2: Inverted Residuals and Linear Bottlenecks," *Conference on Computer Vision and Pattern Recognition,* pp. 4510 - 4520, 2018.

[21] J. Hu, L. Shen, and G. Sun, "Squeeze-and-Excitation Networks," *Conference on Computer Vision and Pattern Recognition,* pp. 7132 - 7141, 2018.

[22] S. Wang, Y. Guan, and L. Shao, "Multi-Granularity Canonical Appearance Pooling for Remote Sensing Scene Classification," *IEEE Transactions on Image Processing,* vol. 29, pp. 5396 - 5407, 2020.

[23] G. Cheng, C. Yang, X. Yao, L. Guo, and J. Han, "When Deep Learning Meets Metric Learning: Remote Sensing Image Scene Classification via Learning Discriminative CNNs," *IEEE Transactions on Geoscience and Remote Sensing,* vol. 56, no. 5, pp. 2811 - 2821, 2018.

[24] I. Loshchilov and F. Hutter, "SGDR: Stochastic Gradient Descent with Warm Restarts," *International Conference on Learning Representations,* 2017.

[25] N. Srivastava, G. Hinton, A. Krizhevsky, I. Sutskever, and R. Salakhutdinov, "Dropout: A Simple Way to Prevent Neural Networks from Overfitting," *Journal of Machine Learning Research,* vol. 15, no. 56, pp. 1929 - 1958, 2014.